\documentclass[10pt,peerreview,cspaper,compsoc]{IEEEtran}
\ifCLASSOPTIONcompsoc
  \usepackage[nocompress]{cite}
\else
\fi
\ifCLASSINFOpdf
   \usepackage[pdftex]{graphicx}
\else
\fi
\usepackage[cmex10]{amsmath}
\usepackage{amsfonts}
\usepackage{color}
\newcommand{\bs}{\boldsymbol}
\newcommand{\cl}{\mathcal}
\newcommand{\tbf}{\textbf}
\newcommand{\squeeze}{\vspace{-2mm}}
\DeclareMathOperator*{\argmin}{\arg\!\min}
\usepackage{soul,xspace}
\newcommand{\ie}{\emph{i.e.}}
\DeclareMathOperator{\Tr}{Tr}
\newcommand{\del}[1]{}

\usepackage{multirow}
\usepackage{xcolor}

\newcommand{\mycolor}[2]
{
	{\color{#1}{#2}}
}

\usepackage{makecell} 
\newcommand{\myline}{\Xhline{2\arrayrulewidth}}

\newcommand{\revise}[1]
{
	{\color{black}{#1}}
}

\newcommand{\mypipe}{\;|\;}

\def\textbestview{}
\def\tuds{TUD Stadtmitte }
\def\tudc{TUD Crossing }

\usepackage{algorithmic}
\usepackage{algorithm}
\usepackage{url}

\begin{document}
%
\title{Discriminative and Efficient Label Propagation on Complementary Graphs for Multi-Object Tracking}
%
%
%
%

\author{Amit~Kumar~K.C.,~\IEEEmembership{Student Member,~IEEE,}
        Laurent~Jacques,
        and~Christophe~De~Vleeschouwer,~\IEEEmembership{Member,~IEEE}
\IEEEcompsocitemizethanks{\IEEEcompsocthanksitem AKC, LJ and CDV are funded by the F.N.R.S. and are affiliated to the ISPGroup, ELEN Department, ICTEAM Institute, Universit\'{e} catholique de Louvain (UCL), B-1348, Louvain-la-Neuve, Belgium.\protect\\
E-mail: \{amit.kc, laurent.jacques, christophe.devleeschouwer\}@uclouvain.be}
\thanks{}}

%
%

\ifCLASSOPTIONpeerreview
\markboth{}
{KC \MakeLowercase{\textit{et al.}}:Discriminative and Efficient Label Propagation on Complementary Graphs for Multi-Object Tracking}
\fi
%


\IEEEcompsoctitleabstractindextext{%
\begin{abstract}
Given a set of detections, detected at each time instant independently, we investigate how to associate them across time. This is done by propagating labels on a set of graphs, each graph capturing how either the spatio-temporal or the appearance cues promote the assignment of identical or distinct labels to a pair of detections. The graph construction is motivated by a locally linear embedding of the detection features. Interestingly, the neighborhood of a node in appearance graph is defined to include all the nodes for which the appearance feature is available (even if they are temporally distant). This gives our framework the uncommon ability to exploit the appearance features that are available only sporadically. Once the graphs have been defined, multi-object tracking is formulated as the problem of finding a label assignment that is consistent with the constraints captured each graph, which results into a difference of convex (DC) program.  We propose to decompose the global objective function into node-wise sub-problems. This not only allows a computationally efficient solution, but also supports an incremental and scalable construction of the graph, thereby making the framework applicable to large graphs and practical tracking scenarios. Moreover, it opens the possibility of parallel implementation.
\end{abstract}

}

\maketitle

\IEEEdisplaynotcompsoctitleabstractindextext

 \ifCLASSOPTIONpeerreview
 \fi
%
\IEEEpeerreviewmaketitle


\section{Introduction}
\IEEEPARstart{I}n this paper, we address the problem of multi-object tracking (MOT). We assume that the targets of interest have been detected at each time instant \cite{delannay2009detection,flueret2008pom, khan2009multiple} and that some appearance features have been extracted. Given this error-prone information, our objective is to link these detections into consistent trajectories using a graph-based formalism. 

Conventionally, a graph-based formalism assigns a node to each detection. Edges are then defined to connect the nodes, and each edge gets a cost that reflects the dissimilarity between the two nodes it connects. Afterwards, a ($K$)-shortest path algorithm \cite{berclaz2011ksp} is typically used to  find the trajectories of the ($K$) targets. Alternatively, other approaches use network flow \cite{globally_optimum_network_flow}, maximum weighted independent set \cite{Brendel11multiobjecttracking}, etc. to solve the same problem. These approaches have been proven to be effective in scenarios for which the features are collected with the same level of accuracy and reliability for each detection. 
With such a stationary measurement process, the likelihood that the detections along a path correspond to the same physical object can be reasonably estimated based on the 
accumulation of dissimilarities (similarities) between (close to)consecutive nodes in the path.
In contrast, these approaches are not appropriate in cases for which appearance features cannot be measured with same accuracy and reliability in every space and time co-ordinates. Such cases are prevalent in many real-life situations. For example, color histograms tend to be noisy in presence of occlusions. In some other cases, highly discriminative features are available only sporadically. This happens, for example, while imaging biological cells under varying illuminations, each illumination level highlighting certain features of the cell. As another example, a digit, printed on the jersey of a player, is available only when it faces the camera. In such cases, the task of tracking multiple objects, while exploiting such features, becomes non-trivial.

Two works have recently addressed this category of problems. 
In their formulation, the authors in \cite{horesh2011global} assume that a discrete set of $L$  possible appearances is known beforehand, which allows the creation of a $L$-layered graph. In the $i$-th layer, running through a node is penalized when the appearance of the node is available and differs from the $i$-th presumed appearance.
Afterwards, a $K$-shortest path algorithm is applied to find the $K$ shortest paths across the $L$-layers. 
This method demonstrates that exploiting sporadic features can significantly improve the tracking performance. However, it is restricted to cases for which the number and the appearance of the targets are known \emph{a priori}.

In contrast, \cite{kc2012iterative} does not make any assumption about the (number of) targets appearances. It proposes a widely applicable iterative hypothesis testing strategy to exploit appearances features that are corrupted by non-stationary noise or are only sporadically available. In short, the authors iteratively consider each node in the graph as a key-node, and investigate how to link this key-node with other nodes in its neighborhood, under the hypothesis that the appearance observed in the key-node position is representative of the actual appearance of the target. 
The method offers the advantage to handle cases for which the discrete set of possible appearances is not known \emph{a priori}. The greedy and iterative nature of the algorithm makes it efficient from a computational and memory usage perspective (no $L$-layered graph). Its main disadvantage is that it is greedy and consequently does not guarantee the global optimality of the solution. 

\begin{figure*}[t]
	\centering
		\includegraphics[width=0.8\linewidth]{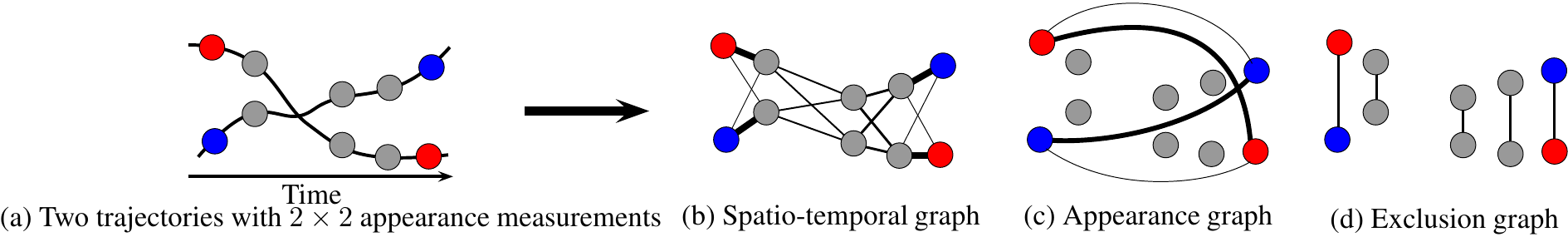}		
   \caption{\textbf{(a)} An example with two targets (red and blue) with associated detections at each time. Gray detections mean that no appearance feature is available. \textbf{(b)} Spatio-temporal graph that depicts the spatio-temporal association between the nodes, \textbf{(c)} Appearance graph that connects nodes even if they are far in time. \textbf{(d)} Exclusion graph in which edges connect nodes that coexist at the same time. 
   {\textbestview} }
	\label{fig:graph_construction}
	\vspace{-5mm}
\end{figure*}

In this paper, we extend our initial contribution in \cite{kc2013label}.
We adopt a graph-based label propagation framework. Therefore, we construct a number of distinct graphs, one for  each appearance feature, apart from the usual spatio-temporal graph. 
Additionally, we also construct an exclusion graph to reflect the fact that two detections that occur at the same time should be assigned to distinct labels.
Hence, we construct $K+2$ `complementary' graphs (one spatio-temporal, $K$ appearance, one exclusion), where $K \ll L$ is the number of appearance features. An example is shown in Figure~\ref{fig:graph_construction}. 
In case of a sport game, for example, the jersey color and the digit, printed on it, can be considered as two appearance features, and result in two distinct appearance graphs.

During graph construction, a node is assigned to each detection. For all the graphs but the exclusivity one, edges connect pairs of nodes with a weight that increases with the similarity between the nodes in terms of space, time or appearance. The higher the weight,  the more likely the two nodes correspond to the same physical target. Exceptionally, the edges of the exclusion graph only connect nodes that cannot belong to the same physical target. This is relevant, for example, when the detections occur at the same time.

Given these graphs, MOT problem is formulated as finding a consistent label assignment, which means that (i) the nodes that are sufficiently close in space/time and/or appearance are labeled similarly, and (ii) the nodes that co-exist at the same time are labeled differently. The consistency of labeling is measured by the labeling energy, which accumulates the difference in the labels between a node and other nodes that are connected to it. Due to the definition of weights in our graph, a good labeling should minimize the energy in the spatio-temporal and the appearance graphs while maximizing the energy due to the exclusion graph. Following our initial contribution in \cite{kc2013label}, our paper formulates the multi-object tracking with sporadic appearance features as a labeling problem in a number of complementary graphs. \revise{In addition to \cite{kc2013label}, it also proposes:
\begin{itemize}
	\item an efficient solution to the labeling problem, splitting the `big' problem into `small' node-wise problems that can be solved locally, optionally based on a parallel implementation (Section~\ref{section:proposed_solution}),
	\item an extension of the local label propagation process to handle incremental/on-line tracking scenarios (Section~\ref{section:incremental_label}).
\end{itemize}
Those two novel contributions make our proposed framework particularly suitable to practical scenarios.
}

The rest of the paper is organized as follows. Section~\ref{section:problem_formulation} formulates the MOT problem as a consistent label assignment problem. Section \ref{section:proposed_solution} proposes the solutions to the label assignment problem. A brief review of the related work is presented in Section~\ref{section:related_work}. Experimental results are presented in Section ~\ref{section:evaluation}. Section~\ref{section:conclusion} concludes our paper.
\vspace{-2mm}
\section{Tracking problem formulation}
\label{section:problem_formulation}
This section first describes the construction of the associated graphs. Afterwards, the multi-object tracking is formulated as a graph-consistent labeling problem.
\subsection{Notation}
Vectors and matrices are denoted with bold lower-case and upper-case symbols respectively while scalar values are denoted by light ones. Upper-case calligraphic letters denote sets. 
\begin{table}[h]
	\centering
	\resizebox{0.99\linewidth}{!}{
	\begin{tabular}{|r p{6.3 cm}|}
		\hline
		$K$ & number of appearance features\\
		$\bs x_i$ & feature vector of the $i$-th sample\\
		$\cl N_i$ & set of neighbors of the $i$-th sample\\
		
		$\bs X^{(i)}$ & features of the neighbors of the $i$-th sample, \emph{i.e.}, \\
		{} & $\bs X^{(i)}:=\cup_{j \in \cl N_i} \{\bs x_j\}$\\
		
		$\bs w_i^\star$ & optimal reconstruction weights for the $i$-th sample\\
		$\cl G$ & a graph of node set $\cl V$, edge set $\cl E$ and weight $\bs W$\\
		$n$ & $=|\cl V|$, number of nodes\\
		
		$\bs L_l^{(+)}$ & Laplacian of the $l$-th graph for $l \in \{0,1,...,K\}$, {$l=0$ for the spatio-temporal graph}\\
		$\bs L^{(-)}$ & Laplacian of the exclusion graph\\
		
		$\bs y_i$ & label distribution assigned to the $i$-th node\\
		$m$ & size of the label vector. \\
		$\bs Y$ & $=(\bs y_1, \bs y_2,..., \bs y_n)^\top$, label assignment matrix\\
		
		$\Delta_d$ & $=\{\bs x \in \mathbb{R}_+^d : \bs 1^\top \bs x=1\}$, probability simplex of a given size $d$\\
		$\cl P_{nm}$ & set of all row-stochastic matrices of size $n\times m$\\
		\hline
		
		\multicolumn{2}{c}{\bf Parameters}\\ \hline
		$\gamma$        & Scaling factor for time (Section \ref{section:graph_construction})                       \\
		$T$             & Connection window size for spatio-temporal graph (Section \ref{section:graph_construction})                        \\
		$v_{\rm max}$   & Maximum speed for gating constraint (Section \ref{section:graph_construction})           \\
		$\alpha_l$      & Weight assigned to the $l$-th labeling energy (Section \ref{section:consistent_labelling}) \\
		$T_c$           & Connection window size for appearance graph (Section \ref{section:incremental_graph_construction})                        \\
		$\sigma$        & `Heat` parameter (Section \ref{section:incremental_graph_construction})                              \\
		$T_o$           & Observation window for bounding complexity (Section \ref{section:label_propagation})   \\
		\hline
	\end{tabular}
	}
	\caption{Notations}
	\label{table:table_of_notations}
	\vspace{-5mm}
\end{table}
\vspace{-5mm}
\subsection{Graph construction}
\label{section:graph_construction}
We consider three distinct types of graphs. Hence, each graph should be constructed separately. Nevertheless, the constructions of spatio-temporal and appearance graphs follow the same approach, derived from the locally linear embedding  (LLE) technique \cite{LLE}. It assumes that data points can be accurately reconstructed by a  weighted linear combination of their local neighbors. We motivate the linearity assumption by the fact that (i) target motion is linear in a small temporal window, and (ii) appearance features lie on a manifold. The number of neighbors is a design parameter, and should be chosen according to the kind of feature and the problem at hand.

In the following, we represent the feature of the $i$-th detection by $\bs x_i$ and that of its neighbors by $\bs X^{(i)}:=\cup_{j \in \cl N_i} \{\bs x_j\}$, where $\cl N_i$ is the set of neighbors of $i$. Afterwards, the graph construction is formulated as the problem of finding the reconstruction weights $\bs w_i^\star$ that minimizes the following optimization problem 
\begin{align}
	\bs w^\star_i=\argmin_{\bs w_i \in \Delta_{|\cl N_i|} } {\|\bs x_i-\bs X^{(i)} \bs 	w_i \|_2^2+\tfrac{\delta}{2}\|\bs w_i\|_2^2},
		\label{eqn:graph_construction}
\end{align}
\revise{where $\Delta_m:=\{\bs w \in \mathbb{R}^{m} \mypipe \bs w \succeq \bs 0, \bs 1^\top \bs w=1\}$ is the probability simplex of a given size $m$. The reason to constrain the weights to belong to the simplex is that it promotes weight vector sparsity. 
To see this, we observe that the simplex constraint is equivalent to enforcing positive weights with unit $\ell_1$-norm, and first consider the case with $\delta=0$ in Equation (\ref{eqn:graph_construction}). When minimizing a quadratic fidelity as the one present in the first term of the cost of Equation (\ref{eqn:graph_construction}) under such $\ell_1$-norm constraint, the solution is generally restricted to a small dimensional facet of the unit $\ell_1$-norm \cite{chen1998atomic,tibshirani1996regression}, \ie, a domain where the solution is sparse.
We favor sparsity as it leads to an efficient optimization in Section~\ref{section:proposed_solution}.} Promoting too much sparsity is however not desired. If a sample is similar to several other samples (\emph{e.g.}, a feature occurs several times along the sequence of detections), the sparse reconstruction selects only one neighbour and ignores the rest. To mitigate this limitation, we add a quadratic part $\tfrac{\delta}{2}\|\bs w_i\|_2^2$, which offers an additional advantage of making the problem strongly convex, resulting in a unique $\bs w_i^\star$. This can be seen as similar to an elastic net regularization in the sense that the sparsity term is imposed by the constraints. \revise{We use $\delta=10^{-2}$.\footnote{Effect of choosing different $\delta$ is discussed in the supplementary material.}}

Once the weights for each data point are computed, we gather them into a graph $\cl G=(\cl V,\cl E,\bs W)$, where
\begin{itemize}
	\item[--]{$\cl V$ is the set of nodes, with $i$-th node corresponding to the $i$-th detection. We denote the number of nodes by $n=|\cl V|$.}
	\item[--] \revise{$\cl E $ defines the connectivity between the samples such that an edge $(i,j)$ is created between nodes $i$ and $j$ only when the weight $\bs w^\star_i(j)$, resulting from Equation~\ref{eqn:graph_construction}, is non-zero, \ie, ${\cl E=\{(i,j) \mypipe \bs w^\star_i(j)>0\}}$}.
	\item[--]{$\bs W$ assigns a weight to each edge such that 
		\begin{align}
			W_{ij}=\begin{cases}
				\bs w_i^\star(j) & \text{ if } (i,j) \in \cl E,\\
				0 & \text{ otherwise}.
			\end{cases}
		\end{align}}
\end{itemize}


Now, we explain the specific issues in the construction of each graph.

\textbf{Spatio-temporal graph}. In case of the spatio-temporal graph, $\bs x_i$ is defined by the time instant $t_i$ and the location information $\bs c_i$ (\emph{e.g.}, bounding box of the detections). Hence, $\bs x_i=(\gamma t_i,\bs c_i)^\top$, where $\gamma$ affects the relative importance of the time difference compared to  the location difference between the data points. \revise{A non-zero $\gamma$ ensures that the prediction of the position of a detection from its neighbors is consistent with both location and time-stamps of the neighbors, assuming that the targets move at constant velocity in a small temporal neighborhood. We use $\gamma=3$ pixels/frame. Our experiments (Figure \ref{fig:effect_of_parameters}) show that this choice has little impact on the performance.}  

The neighbors $\cl N_i$ are defined to be the samples whose time indices fall within a small temporal window of size $T$ \revise{without falling under the gating constraint defined below to build the exclusion graph. $T$ should be large enough to bridge local missed detections, but also small enough so that linear motion assumption holds.} 
We use $T=10$ frames. 

\textbf{Appearance graph}. In case of the appearance graph, $\bs x_i$  corresponds to an appearance feature (\emph{e.g.}, color histograms, etc.). Since we are considering the fact that a feature might occur only sporadically, $\cl N_i$ is defined to constitute all the samples except the samples that co-occur with the $i$-th sample and that do not have appearance features. 

\textbf{Exclusion graph}. This graph captures the constraints associated to the fact that some detections cannot share the same labels. For example, two detections that occur at the same time instant should have different labels. This is usually referred to as \emph{time exclusivity}. This information is encoded by setting $W_{ij}=1$ if $t_i=t_j$. In addition, we can enforce the spatial constraint such that a target cannot have a large spatial displacement over short time interval. We encode this \emph{gating} constraint by setting $W_{ij}=1$ if $\|\bs c_i-\bs c_j\|_2>v_{\text{max}} |t_i-t_j|$, where $v_{\text{max}}$ is the maximum speed of the target. Thus, $\cl N_i$ comprises of the detections that either co-exist with the $i$-th detection or violate the gating constraint. 
\vspace{-5mm}
\subsection{Multi-object tracking as consistent labeling problem}
\label{section:consistent_labelling}
Given a set $\cl V$ of $n$ vertices (\emph{i.e.}, the detections or the \emph{tracklets} in tracking scenario), we consider that a label assignment $\bs Y=(\bs y_1, ..., \bs y_{n})^\top$ is defined to assign a $m$-dimensional\footnote{\revise{Ideally, $m$ should be equal to the number of targets plus one (for false positive). Since, in general, we do not know the number of targets \emph{a priori}, we set $m=n$, considering the worst case in which each detection corresponds to a different target.}} label distribution $\bs y_i \in \Delta_m$ to the $i$-th node, where $\Delta_m$ 
is the $m$-dimensional probability simplex. \revise{Each dimension of the label distribution $\bs y_i$ corresponds to a target. Formally, the $k$-th dimension, $\bs y_i(k), \; k=1,\cdots,m$, can be interpreted as the probability of the node $i$ being the $k$-th target.} Consequently, $\bs Y$  is a row-stochastic matrix, with each row summing to unity. Therefore, we write $\bs Y \in \cl P_{nm}$, where $\cl P_{nm}$ is the set of all row-stochastic matrices of size $n \times m$. We consider a graph $\cl G=(\cl V,\cl E,\bs W)$ as explained earlier. This graph is assumed to assign large positive weights to edges that connect vertices that are likely to have similar labels (typically because they are close in time and space, or because they have similar appearance). In \cite{zhu2003semi}, a harmonic function approach is introduced to measure the inconsistency of the label assignment matrix $\bs Y$ with respect to the graph $\cl G$. Specifically, it measures the $\ell_2$-norm of the difference between the labels assigned to nodes that are connected in the graph $\cl G$, and the labeling energy, also known as the harmonic energy \cite{zhu2003semi}, is defined as
%
\begin{align}
	E_{\bs L}(\bs Y):=\tfrac{1}{2}\sum_{i=1}^{n}{\sum_{j =1}^n{W_{ij} \| \bs y_i-\bs y_j\|_2^2}} = \textbf{Tr}(\bs Y^\top \bs L \bs Y),
	\label{eqn:general_energy}
	\vspace{-2mm}
\end{align}
where $\textbf{Tr}$ is the trace of a matrix and $\bs L$ is the graph Laplacian, defined as $\bs L=\bs D-\bs W$, where $\bs D$ is a diagonal matrix whose $i$-th diagonal element is $D_{ii}:= \sum_{j \in {\cl N}_i}{W_{ij}}$. \revise{Due to the definition of weights in our graphs, we have $D_{ii}=1$. Therefore, $\bs D$ is an identity matrix.}
For a graph with non-negative weights, \emph{i.e.}, $W_{ij} \geq 0$, $\bs L$ is positive semi-definite and consequently the labeling energy in Equation~(\ref{eqn:general_energy}) is convex in $\bs Y$. 

In our framework, we have $K+2$ distinct graphs. As all the graphs have the same set of nodes, we frequently refer to a graph by its Laplacian $\bs L$ in the sequel. We represent the exclusion graph by $\bs L^{(-)}$, and other graphs by $\bs L_l^{(+)}, l \in \{0,...,K\}$, where $l=0$ corresponds to the spatio-temporal graph and $1 \leq l \leq K$ corresponds to the $l$-the appearance graph. We explicitly introduce the minus (respectively, plus) superscript to emphasize that we would like to maximize (respectively, minimize) the labeling energy on the corresponding graph.

Given the measure of labeling energy on each graph, we want to define a label assignment $\bs Y^\star$ that minimizes the labeling energies due to $\bs L_l^{(+)}$ and maximizes the labeling energy due to $\bs L^{(-)}$. 
Mathematically, we have
\squeeze
\begin{align}
	\bs Y^\star&:= \argmin_{\bs Y \in \cl P_{nm}}\sum_{\revise{l=0}}^{K}{\alpha_l E_{\bs L_l^{(+)}}(\bs Y)}-E_{\bs L^{(-)}}(\bs Y) \nonumber \\
	&=\argmin_{\bs Y \in \cl P_{nm}} E_{\bs L_{\rm eff}^{(+)}}(\bs Y)-E_{\bs L^{(-)}}(\bs Y) 
	\label{equation:trace_ratio}
\end{align}
where $\bs L^{(+)}_{\rm eff}:=\sum_{l=0}^{K}{\alpha_l \bs L_p^{(+)}}$, and $\alpha_l \geq 0$ weighs the contribution of labeling energy due to $l$-th graph. \revise{The choice of $\alpha_l$ depends on the scenario at hand, \ie, on the prior knowledge available about the relevance of the features. For example, while tracking sport players, the decrease in labeling energy associated to the color graph is not of primary importance since the players from the same team have similar colors. Hence, detections sharing the same color might correspond to distinct players/labels. In such case, it is meaningful to lower the weight assigned to the color graph as compared to the spatio-temporal graph.
In other cases, for which a unique specific color is assigned to each target, a large weight should be assigned to the color graph to force the assignment of distinct labels to detections having different colors. }
Since $\alpha_l \geq 0$ and $\bs L_l^{(+)}$ is positive semi-definite, $\bs L^{(+)}_{\rm eff}$ is also positive semi-definite.
Given $\bs Y^\star$, the $i$-th node is assigned the label that corresponds to the largest entry in $\bs y_i^\star$. 

\section{Graph-consistent labels computation}
\label{section:proposed_solution}
In this section, we explain how to compute the solution $\bs Y^\star$ of the problem, defined in Equation~(\ref{equation:trace_ratio}). First, we present a global label assignment solution, based on the difference of convex programming. Afterwards, we introduce a node-wise optimization approach to solve the problem efficiently.
\vspace{-1mm}
\subsection{Joint label assignment optimization}
\label{section:joint_optimization}
 
Let us rewrite Equation~(\ref{equation:trace_ratio})
as
\begin{align}
	\bs Y^\star&=\argmin_{\bs Y \in \cl P} \tbf{Tr}(\bs Y^\top \bs L^{(+)}_{\rm eff} \bs Y)-\tbf{Tr}(\bs Y^\top \bs L^{(-)}\bs Y) \nonumber \\
	&:= \argmin_{\bs Y \in \cl P} \bigl[g(\bs Y):=f(\bs Y)-h(\bs Y)\bigr]
	\label{eqn:main_equation}
\end{align}

As $\bs L^{(+)}_{\rm eff}$ and $\bs L^{(-)}$ are positive semi-definite matrices, both $f(\bs Y):=\tbf{Tr}(\bs Y^\top \bs L^{(+)}_{\rm eff} \bs Y)$ and $h(\bs Y):=\tbf{Tr}(\bs Y^\top \bs L^{(-)} \bs Y)$ are convex in $\bs Y$, whereas $g(\bs Y)$ is non-convex. Specifically, Equation~(\ref{eqn:main_equation}) belongs to a family of \emph{difference of convex (DC)} problems, and an iterative majorization-minimization algorithm can be used to solve the problem \cite{bharat2009convergence}, as presented in Algorithm~\ref{algorithm:global_algorithm}. Starting with a random label distribution $\bs Y^{(1)} \in \cl P$, the algorithm iteratively linearizes $h(\bs Y)$ around the $k$-th iterate $\bs Y^{(k)}$ and solves the resulting convex function $f(\bs Y)-\nabla h^\top\left(\bs Y^{(k)}\right) \bs Y$ using the projected gradient method \cite{projected_gradient_method}. The number of iterations $T_{\text{joint}}$ depends on the convergence tolerance.

\begin{algorithm}[h]
	\footnotesize
	\caption{Joint label assignment optimization}
	\label{algorithm:global_algorithm}
	\begin{algorithmic}
		\STATE{\textbf{Input}}
		\STATE{$\quad $ Graph Laplacians: $\{\bs L_l^{(+)}, l=0,...,K\}, \bs L^{(-)}$}
		\STATE{$\quad $ Scaling weights: $\{\alpha_l, l=0,...,K\}$}
		\STATE{$\quad $ Number of iterations: $T_{\text{joint}}$}
		
		\STATE{\textbf{Output}}
		\STATE{$\quad $ Label assignment matrix: $\bs Y^\star$}
		
		\STATE{\textbf{Procedure:}}				
		\STATE{$\quad $ Choose an initial solution $\bs Y^{(1)} \in \cl P_{nm}$ randomly.} 
		\STATE{$\quad $ \textbf{For} $k=1,...,T_{\text{joint}}$}
			\STATE{$\quad \quad $ Compute $\nabla h(\bs Y^{(k)})$, gradient of $h(\bs Y)$ at $\bs Y^{(k)}$.}
			\STATE{$\quad \quad $ Solve the convex optimization problem}
			\STATE{$\quad \quad \quad$  $\bs Y^{(k+1)} \leftarrow \argmin\limits_{\bs Y \in \cl P_{nm}} \bigl[f(\bs Y)-\nabla h^\top(\bs Y^{(k)}) \bs Y\bigr]$}
			\STATE{$\quad  \quad \quad $ by the projected gradient method \cite{projected_gradient_method}.}
		\STATE{$\quad $ \textbf{End For}}
		\STATE{$\quad $ \textbf{Return} $\bs Y^\star \leftarrow \bs Y^{(T_{\text{joint}})}$.}
	\end{algorithmic}
\end{algorithm}

\revise{It is worth noting that the gradient of $\tbf{Tr}(\bs Y^\top \bs L \bs Y)$ is $(\bs L +\bs L^\top)\bs Y$. Therefore, both $\bs L$ and its transpose $\bs L^\top$ are considered identically during gradient descent.}

\textbf{Complexity analysis:}
Since there are $n$ nodes, the graph Laplacian is a $n \times n$ matrix. Each node is assigned to a $m$-dimensional label distribution. Consequently, $\bs Y$ is a $n \times m$ matrix. 
The projected gradient method \cite{projected_gradient_method} performs gradient descent step followed by projection step for $T_p$ times. Each step has a naive complexity of $\cl O(n^2m)$, which can be improved to $\cl O(kmn)$ if the graph Laplacian is $k$-sparse. Thus, the overall complexity is $\cl O(n^2m T_p T_{\text{joint}})$. The parameters $T_p$ and $T_{\text{joint}}$ depend on a fixed tolerance value.

The main disadvantage of the above solution is that its computational complexity grows quadratically with the number of nodes. Therefore, it cannot scale to large graphs. Furthermore, it can only handle off-line tracking problems because the optimization problem formulation is based on the whole graph.

In the sequels, we describe how to circumvent these limitations based on a node-wise decomposition.
\vspace{-5mm}
\subsection{Node-wise label assignment optimization}
\label{section:efficient_label_propagation}
To address  the complexity issue of the joint label propagation algorithm, we adopt a node-wise decomposition of the objective function. That is, instead of solving a ``big'' and ``global'' optimization problem, each node updates locally and sequentially its label distribution to decrease the global objective.
The approach is similar to the Gauss-Seidel iteration (or, co-ordinate descent approach). The advantages of such decomposition are twofold. First, the computational complexity gets significantly reduced, making the framework applicable to large graphs, potentially based on parallel implementation. Second, as we solve the problem by iterating over the nodes, it becomes possible to handle tracking problems for which the graphs grow incrementally, as new detections are gradually computed along the time. 

In the remainder of the section, we first explain our proposed efficient and node-wise label propagation solution, and derive the conditions under which the global objective function monotonically decreases. Afterwards, we introduce a strategy to scale up the algorithm using parallel implementation. 
\subsubsection{Node-wise decomposition}
In this section, we first generalize the energy in Equation~(\ref{eqn:general_energy}) by replacing the $\frac{1}{2}\|\bs y_i-\bs y_j\|_2^2$ term by a convex and symmetric function $\phi(\bs y_i,\bs y_j)$. Afterwards, we decompose the global optimization problem in Equation~(\ref{eqn:main_equation}) into a node-wise optimization problem such that the high dimensional optimization problem is turned into a sequence of small problems in each node. In doing so, we derive the class of $\phi$ functions that guarantees monotonic decrease of the objective function.\footnote{Detailed derivation is provided in the supplementary material.}


Formally, replacing the $\ell_2$-norm by $\phi$ in Equation~(\ref{eqn:general_energy}), we write the objective function in Equation~(\ref{eqn:main_equation}) as 
\begin{align}
g(\bs Y)&=\sum_{i=1}^{n} \sum_{j=1}^n \left[\sum_{l=0}^K\alpha_l W_{ij}^{(l)}-W_{ij}^{(-)}\right] \phi(\bs y_i,\bs y_j) \nonumber \\
		&\equiv \sum_{i=1}^{n} \sum_{j=1}^{n} W_{ij}^{(\rm eff)} \phi(\bs y_i,\bs y_j), \label{eqn:def_weff} 
\end{align}
where we define $W_{ij}^{(\rm eff)}:=\sum_{l=0}^K \alpha_l W_{ij}^{(l)}-W_{ij}^{(-)}$.
Denoting $\widetilde A_{ij} := A_{ij} + A_{ji}$, we then isolate the contribution of the $p$-th node as
\begin{align}
\!\!g(\bs Y)&=\sum_{j} W_{pj}^{(\rm eff)} \phi(\bs y_p,\bs y_j)+ \sum_{i \neq p}\sum_j W_{ij}^{(\rm eff)} \phi(\bs y_i,\bs y_j) \nonumber \\
		&= \sum_j \widetilde W_{pj}^{(\rm eff)}\,\phi(\bs y_p,\bs y_j) + \sum_{i \neq p} \sum_{j \neq p} W_{ij}^{(\rm eff)} \phi(\bs y_i,\bs y_j)\label{eqn:main_equation1}\\
		&= g_p(\bs y_1,\cdots,\bs y_n)+\sum_{i \neq p} \sum_{j \neq p} W_{ij}^{(\rm eff)} \phi(\bs y_i,\bs y_j) \label{eqn:def_gi}	
\end{align}
where we assume $\phi(\bs y_i,\bs y_i)=0$ and $\phi(\bs y_i,\bs y_j)=\phi(\bs y_j,\bs y_i)$ in Equation~(\ref{eqn:main_equation1}), and we introduce $g_p(\bs y_1,\cdots,\bs y_n):=\sum_j \widetilde W_{pj}^{(\rm eff)}\phi(\bs y_p,\bs y_j) $ for brevity in Equation~(\ref{eqn:def_gi}).

Given $\bs Y^{(k)}=(\bs y_1^{(k)},\cdots,\bs y_n^{(k)})^\top \in \cl P_{nm}$, we choose an index $p \in \{1,\cdots,n\}$ and compute a new iterate $\bs Y^{(k+1)}=(\bs y_1^{(k+1)},\cdots,\bs y_n^{(k+1)})^\top \in \cl P_{nm}$ that satisfies 
\begin{align}
	\bs y_{i}^{(k+1)} \begin{cases}
						=\bs y_{i}^{(k)} & \text{if } i \neq p,\\
						\in\argmin\limits_{\bs y \in \Delta_m} g_i(\bs y_1^{(k)},\cdots,\bs y,\cdots,\bs y_n^{(k)}) & \text{if } i=p.
						\end{cases}
\end{align}
Then, by construction,
\begin{align*}
g(\bs Y^{(k+1)})& =g_p(\bs Y^{(k+1)}) +\sum_{i \neq p} \sum_{j \neq p} W_{ij}^{(\rm eff)} \phi(\bs y_i^{(k)},\bs y_j^{(k)})\\
&\leq  g_p(\bs Y^{(k)}) +\sum_{i \neq p} \sum_{j \neq p} W_{ij}^{(\rm eff)} \phi(\bs y_i^{(k)},\bs y_j^{(k)})\\
	&=g(\bs Y^{(k)})
\end{align*}
Therefore, we conclude that under the following assumptions:
\begin{itemize}
	\item the loss function $\phi(\cdot, \cdot)$ is convex,
	\item the loss function is \emph{coincident}\footnote{\revise{The coincidence property will make the loops irrelevant and generally we do not need loops in the graph.}}, \ie, $\phi(\bs y_i,\bs y_i)=0$,
	\item and the loss function is \emph{symmetric} with respect to its arguments, \emph{i.e.}, $\phi(\bs y_i,\bs y_j)=\phi(\bs y_j,\bs y_i)$,
\end{itemize}		
	the optimization step at any fixed node $p$
	\begin{align}
		\bs y_{p}^{(k+1)} &\in \argmin_{\bs y \in \Delta_m} g_p(\bs y_1^{(k)},\cdots,\bs y,\cdots,\bs y_n^{(k)}) \nonumber \\
									&=\argmin_{\bs y \in \Delta_m} \sum_j \widetilde W_{pj}^{(\rm eff)}\,\phi(\bs y,\bs y_j^{(k)})
					  \label{eqn:def_xi_k1_again}
	\end{align}	
monotonically decreases the objective function $g(\bs Y)$. Equation~(\ref{eqn:def_xi_k1_again}) is still a DC problem and it can be solved by using \emph{majorization-minimization} technique, as discussed in Section~\ref{section:joint_optimization}. \revise{It has to be noted that when $\phi$ is chosen to be the $\ell_2$-norm, the above conditions are satisfied.}

The \emph{label propagation} process is finally achieved by sequentially updating the label distribution over the nodes, possibly $T_{\text{con}} > 1$ times, until $g(\bs Y^{(k)})$ does not decrease any more. We summarize the overall process in Algorithm~\ref{algorithm:node_wise_coordinate_descent}. Note that we do not assume anything about the structure of the graph, thereby allowing loops in the graph. 

\begin{algorithm}[h]
	\caption{Node-wise label assignment algorithm}
	\footnotesize
	\begin{algorithmic}
		\STATE{\textbf{Input}}
			\STATE{$\quad$ Weight matrices: $\{\bs W^{(l)}, l=0,\cdots K\}, \bs W^{(-)}$}
			\STATE{$\quad$ Scaling weights: $\{\alpha_l, l=0,\cdots K\}$}
			\STATE{$\quad$ Number of iterations: $T_{\rm con}$}
		\STATE{\textbf{Output}}
			\STATE{$\quad$ Label assignment matrix: $\bs Y^\star$}
		\STATE{\textbf{Procedure}}
			\STATE{$\quad$ Set $\bs W^{(\rm eff)} \leftarrow \sum_l \alpha_l \bs W^{(l)}-\bs W^{(-)}$}
			\STATE{$\quad$ Set $\widetilde{\bs W}^{(\rm eff)} \leftarrow \bs W^{(\rm eff)}+\bs W^{(\rm eff)\top}$}
			\STATE{$\quad$ Choose initial solution, $\bs Y^{(1)} \in \cl P_{nm}$}
			\STATE{$\quad$ Set $k \leftarrow 1$}
			\STATE{$\quad$ \textbf{For} $t=1,\cdots, T_{\text{con}}$}
				\STATE{$\quad$ $\quad$ Initialize $\cl U \leftarrow \cl V$}		
				\STATE{$\quad$ $\quad$ \textbf{While} $\cl U \neq \emptyset$}		
					\STATE{$\quad$ $\quad \quad$ Select a node $p$ from $\cl U$}
					\STATE{\revise{$\quad$ $\quad \quad$ Solve $\tilde{\bs y} \leftarrow \argmin\limits_{\bs y \in \Delta_m} \sum_j \widetilde W_{pj}^{(\rm eff)}\,\phi(\bs y,\bs y_j^{(k)})$}}
					\STATE{\revise{$\quad$ $\quad \quad$ $ \bs Y^{(k+1)} \leftarrow(\bs y_1^{(k)},\cdots,\bs y_{p-1}^{(k)}, \tilde{\bs y},\bs y_{p+1}^{(k)},\cdots,\bs y_n^{(k)})^\top$}}
					\STATE{$\quad$ $\quad \quad$ $ \cl U\leftarrow \cl U \setminus \{p\}$}
					\STATE{$\quad$ $\quad \quad$ $ k \leftarrow k+1$}					
				\STATE{$\quad$ $\quad $ \textbf{End While}}
			\STATE{$\quad$ \textbf{End For}}
			\STATE{$\quad$ \textbf{Return} $\bs Y^\star \leftarrow \bs Y^{(T_{\text{con}})}$}
	\end{algorithmic}
	
	\vspace{1mm}
	\revise{\underline{Note:} we have observed that the order in which $p$ is chosen from $\cl U$ does not affect the labeling energy much. Consequently, we chose nodes in the sequential order.} 
	
	\label{algorithm:node_wise_coordinate_descent}
\end{algorithm}

\textbf{Complexity analysis:} Each node solves a $m$-dimensional DC program using the projected gradient method. Let the number of iterations required for the convergence of the projected gradient method be $T_{p^\prime}$, which is comparable to $T_p$ in Section~\ref{section:joint_optimization}. 
The complexity of the DC optimization in a specific node is therefore $\cl O(m T_{p^\prime})$. Since there are $n$ nodes and since we traverse the nodes $T_{\text{con}}$ times, the overall complexity is $\cl O(mnT_{p^\prime}T_{\text{con}})$. From experiments, we have seen that $T_{\text{con}} \ll T_{\text{joint}}$. Comparing with the complexity of joint approach, which is $\cl O(n^2m T_p T_{\text{joint}})$, the node-wise decomposition approach has an improvement of $\cl O(n T_{\text{joint}}/T_{\text{con}})$, which becomes significant as $n$ increases, making it a better choice for large-scale problems as confirmed by our experiments.
\vspace{-2mm}
\subsubsection{Parallel implementation}
\label{section:parallel_implementation}
The node-wise decomposition of the objective function also paves the way for a parallel implementation of the label optimization. This allows our proposed approach to scale up further with the size of the graph. In this section, we first derive a condition under which the parallelization of the coordinate descent decreases the objective function. 

We denote the set of nodes for parallel descent by $\cl J$ and its complement by $\bar{\cl J}:=\cl V \setminus \cl J$.
Then, we decompose the objective function as \footnote{Detailed derivation is provided in the supplementary material.}
\begin{align}
g(\bs Y)&=\sum_{i \in \cl J} g_i(\bs Y)-\sum_{i \in \cl J} \sum_{j \in \cl J} \widetilde{W}_{ij}^{(\rm eff)}\phi(\bs y_i,\bs y_j)\nonumber \\
		& \quad{} +\sum_{i \in \bar{\cl J}} \sum_{j \in \bar{\cl J}} W_{ij}^{(\rm eff)}\phi(\bs y_i,\bs y_j)\label{eqn:general_decomposition}
\end{align}

The negative terms in Equation~(\ref{eqn:general_decomposition}) are called \emph{interference} terms. To nullify these terms, we pickup the nodes in $\cl J$ such that there are no edges between them, \emph{i.e.}, $\forall (i,j) \in \cl J \times \cl J, \widetilde{W}^{(\rm eff)}_{ij}=0$. Under this condition, we can write
\begin{align}
	g(\bs Y)&= \sum_{i \in \cl J} g_i(\bs Y)+\sum_{i \in \bar{\cl J}} \sum_{j \in \bar{\cl J}} W^{(\rm eff)}_{ij}\phi(\bs y_i,\bs y_j)\label{eqn:general_decomposition_parallel}
\end{align}
and solve the local optimization problem 
 \begin{align}
 	\bs y_i^{(k+1)} & \in \argmin_{\bs y \in \Delta_m} g_i(\bs y_1^{(k)},\cdots,\bs y,\cdots,\bs y_n^{(k)})
 	\label{eqn:def_xi_k1}
 \end{align}
in parallel for each node $i \in \cl J$. 
Then, the resulting label assignment matrix $\bs Y^{(k+1)}$, defined as
	\begin{align}
	\bs y_{i}^{(k+1)} \begin{cases}
						\in\argmin\limits_{\bs y \in \Delta_m} g_i(\bs y_1^{(k)},\cdots,\bs y,\cdots,\bs y_n^{(k)}) & \text{if } i \in \cl J,\\
						=\bs y_{i}^{(k)} & \text{otherwise},	\nonumber				
						\end{cases}
\end{align}
decreases monotonically the objective function, \ie, $g(\bs Y^{(k+1)}) \leq g(\bs Y^{(k)})$. 
As a consequence, as long as the nodes that are processed in parallel \revise{are not neighbors,} a monotonic decrement of the objective function is guaranteed. In Section~\ref{section:results}, we demonstrate the benefit of parallelization with a simple yet effective batch-based scheduling approach.
\vspace{-2mm}
\section{From off-line to incremental label propagation}
\label{section:incremental_label}
In previous sections, we described the off-line graph construction and label propagation steps. 
However, in many real-life applications, detections arrive progressively along the time. To handle such scenarios, while being as close as possible to the off-line formalism, we embed the node-wise label propagation within an incremental graph construction process. Once the novel detections arrive, the graph is incremented by incorporating them. Afterwards, we re-optimize the label distribution by iterating over the nodes using the node-wise decomposition. 

In the incremental graph construction, we do not have access to the future samples. Consequently, the LLE-based graph construction of Section \ref{section:graph_construction} cannot be used. This has two implications. First, we need to define an explicit strategy to gradually incorporate new targets in the scene. 
Second, the implicit linear motion model cannot be embedded while constructing the spatio-temporal graph since future detection locations are not known at construction time. 

The remainder of the section first explains how new detections are connected to the existing nodes. It then describes how labels are propagated through the incremented graph.

\vspace{-4mm}
\subsection{Incremental Graph Construction}
\label{section:incremental_graph_construction}
We assume that the detections arrive sequentially along the time. Let the set of detections at time $t$ be denoted by $\cl D^{(t)}$. Also, let the graph up to time $t-1$ be $\cl G^{(t-1)}=(\cl V^{(t-1)},\cl E^{(t-1)},\bs W^{(t-1)})$. Since the graph evolves with time, we denote the number of nodes and the size of the label vector explicitly by $n^{(t)}$ and $m^{(t)}$ respectively.

The incrementation differs depending on whether we are dealing with
the spatio-temporal graph, the appearance graph(s) or the exclusion graph.
In all graphs, the new detections are first added to the set of vertices $\cl V^{(t-1)}$ to generate $\cl V^{(t)}$. Edges and weights are incremented separately for each graph as follows: 

\textbf{Exclusion graph.} We create new edges between the nodes that occur at time $t$. 
Also, we create edges from the  nodes at time $t$ to the existing previous nodes if they are not within the \emph{gating region}. Each exclusion edge has a weight 1.

\textbf{Spatio-temporal and appearance graphs.} We connect each node at time $t$ with the nodes in a window $[t-T_{c},t]$, where $T_{c}$ is the connection window size. Large $T_c$ results in dense graphs whereas small $T_c$ results in sparse graphs. Once the neighborhood is defined, we assign a weight $W_{ij}$ between a novel node $i$ and an existing node $j$ as
\begin{equation}
  W_{ij}=\left\{\begin{array}{ll}
      \exp(-\tfrac{1}{\sigma^2} d(\bs x_i,\bs x_j)^2) & \text{if } |t_i-t_j| \leq T_c, \\
      0 & \text{otherwise,}			
    \end{array}\right.
  \label{eqn:weight_computation}
\end{equation}
where $t_i$ and $\bs x_i$ denote the time instant and the features of the $i$-th node respectively, $d(\cdot,\cdot)$ measures the dissimilarity between the features $\bs x_i$ and $\bs x_j$, 
and $\sigma$ is a scaling parameter. $T_c$ and $\sigma$ parameters are adapted to each kind of graph.
In our experiments, $T_c$ is set to 10 frames for the spatio-temporal graph (\revise{as in off-line graph construction}), but is extended up to 200 frames in the appearance graph to bridge the gaps caused by the sporadic nature of the feature. 
\revise{The parameter $\sigma$ should be larger than the typical distance measured between the features of two detections corresponding to the same targets, while being smaller than the typical distance measured between distinct targets. In practice, our values for $\sigma$ have been selected by looking at the two distributions of distances between pairs of detections that correspond to the same/different targets.\footnote{These distributions should ideally be derived from ground-truth data. When such a ground-truth is not available, we shwon in the supplementary material that reasonable $\sigma$ can simply be inferred by comparing two distributions of distances measured between either neighboring or co-existing detections. Another alternative consists in building on reliable tracklets to identify pairs of detection corresponding to similar/distinct targets.}
 Specifically, we use $\sigma=20$ in the spatio-temporal graph and $\sigma=0.05$ in the appearance graph. Also, we use $d(\cdot,\cdot):=\|\cdot - \cdot\|_2$ but any other distance measure can be envisioned.} 

To account for the cases in which some detections (nodes) are likely to correspond to new targets, we introduce a virtual source node in the graph. This source node  is connected to every node in the spatio-temporal graph. The weight of the edge connecting the source node to the $i$-th node is represented by $w^{(s)}_i$. This weight depends on the prior knowledge we might have about where and/or when a target is likely to appear in the field of view. \revise{In our case, we consider that a new target appears either in the beginning of the tracking process, or when entering the scene on the borders of the image. Therefore, the weights should be large for the detections that are close to the image border and/or that appear in the beginning of the tracking.} For the $i$-th detection, we compute the smallest distance $d_i^{(\min)}$ from the detection to the image border. Then, we compute $w^{(s)}_i$ by replacing $d(\cdot,\cdot)$ by $d_i^{(\min)}$ in Equation~(\ref{eqn:weight_computation}). Note that when some prior knowledge is available about the appearance of the targets entering the scene, \emph{e.g.}, because the digit of the players sitting on the dug-out in team sport games is known, edges to the source node could be defined in the appearance graph as well.
\revise{Once the weights are defined, they are normalized such that $\sum_{\bs j \in \{\cl N_i \cup s\}} W_{ij}=1$.} 
\vspace{-3mm}
\subsection{Label propagation in the incremented graph}
\label{section:label_propagation}
After incrementing the graphs, we perform node-wise label propagation. 
We denote the labels distribution over $\cl G^{(t)}$ after $k$ iterations of the label propagation process by $\bs Y^{(t,k)}$. Moreover, $\bs Y^{(t)}$ denotes the labels distribution after the convergence of the propagation process at time $t$. We  first initialize the label distribution matrix at time $t$, denoted by \revise{$\bs Y^{(t,1)}$, by augmenting the label distribution matrix at time $t-1$, denoted by a $n^{(t-1)} \times m^{(t-1)}$-dimensional matrix $\bs Y^{(t-1)}$, as follows:
\begin{align}
	\bs Y^{(t,1)}&=\begin{pmatrix} \bs Y^{(t-1)} \: | \:  \bs 0_{n^{(t-1)} \times |\cl D^{(t)}|} \\[1mm] \bs U^{(t)}\end{pmatrix}
	\label{eqn:label_augmentation}
\end{align}
where $\bs 0_{n^{(t-1)} \times |\cl D^{(t)}|}$ is a $n^{(t-1)} \times |\cl D^{(t)}|$-dimensional zero matrix and $\bs U^{(t)}$  is a $|\cl D^{(t)}| \times \left(|\cl D^{(t)}|+m^{(t-1)}\right)$-dimensional matrix such that $U^{(t)}_{ij}=1/(|\cl D^{(t)}|+m^{(t-1)})$. Obviously, $\bs U^{(t)}$ is a (uniform) row-stochastic matrix, and a uniform label distribution is assigned to the novel nodes.}


After initialization, we iterate over all the nodes (except the virtual source node) and solve the node-wise optimization problem, introduced in Section~\ref{section:efficient_label_propagation},
\begin{align}		
		\bs y_{i}^{(t,k+1)} \in \argmin_{\bs y \in \Delta_{\revise{m^{(t)}}}} & \, g_i(\bs y_1^{(t,k)},\cdots,\bs y,\cdots,\bs y_{m^{(t)}}^{(t,k)}) \nonumber \\[-10pt]
		& \enspace + w_i^{(s)} \phi(\bs y,\bs e_i),
		\label{eqn:incremental_label_propagation}
\end{align}	

where $\bs e_i \in \Delta_{\revise{m^{(t)}}}$ is a singleton vector having 1 at the $i$-th index and zero \revise{elsewhere}. It promotes the assignment of a new label to the $i$-th node when $w_i^{(s)} \approx 1$. 

To bound the complexity of our incremental framework, and to turn it into an on-line procedure, we consider a sliding window $[t-T_o,t]$ and forget the history of the graph outside the window. Afterwards, the distributions of nodes that lie outside the window are frozen, and the node-wise optimization, defined in Equation~(\ref{eqn:incremental_label_propagation}), is only considered for the nodes that belong to the window. The window size $T_o$ trades-off the tracking accuracy and the computational (and memory) resources. 

\vspace{-2mm}
\section{Related work}
\label{section:related_work}
This section provides a brief review of the recent and related works under the following categories:

\textbf{Label propagation in graphs}. Propagation of labels in a graph is often used in semi-supervised learning approaches, and a concise survey of recent developments in this field can be found in \cite{robust_and_scalable_graph} and references therein. In short, most of these approaches assume that the label of a node is approximated as the linear combination of the labels of its neighbours \cite{Wang06labelpropagation}. In \cite{mixed_label_propagation}, the authors use a mixed label propagation in which (i) they measure the bipolar similarity (\emph{e.g.}, Karl Pearson's correlation coefficient that lies in the range [-1,1]) between the samples, and (ii) construct a `positive' and a `negative' graphs based on the sign of the coefficient. Afterwards, they minimize the ratio between labeling energies due to the positive and negative graphs. This is done by semi-definite relaxation to assign a binary label to each node of the graph. Our method differs from \cite{mixed_label_propagation} both in the definition of the graph similarities, and the label propagation method. Specifically, since we use multi-class labels instead of binary labels, and impose that the label distribution at each node should lie on a probability simplex, our problem is difficult to cast into their formalism. 

\textbf{Message passing}. Message passing (belief propagation) approaches have been used to label the nodes in a graph in tracking/recognition \cite{lu_identification, kc2012prioritizing}, image completion scenario \cite{felzenszwalb2006efficient}, etc. Each node gathers messages from its neighbors, optimizes locally a problem, and then transmits its message. This approach has been shown to be exact in trees but the convergence is not guaranteed in presence of loops. In contrast, we do not assume any structure of the graph to guarantee the convergence of our approach.

In \cite{lu_identification}, a subset of the nodes are initially labelled and then a CRF is used to infer the label of the remaining nodes. For this, the  authors compute various appearance features and assume that the features are always available with similar accuracies. Hence, their approach cannot exploit appearance features that are sporadic or affected by non-stationary noise. In \cite{kc2012prioritizing}, the authors utilized such non-stationary and sporadic features to prioritize the propagation of belief related to the label probability distribution. Even though this approach exploits sporadically available appearance features, it relies on the assumption that the target appearances are known beforehand, which is not the case of our approach. 


{\tbf{Mutual exclusion.} Mutual exclusion has been considered in \cite{online_discriminative, how_does_identity_recognition} to learn discriminative appearance features. 
In these papers, first of all, a low-level but reliable tracker is used to connect unambiguous detections into tracklets. Afterwards, positive samples are defined by pairs of detections that belong to the same tracklet, while negative samples correspond to pairs that belong to tracklets that likely correspond to distinct objects (because they overlap in time). Lastly, these samples are used to train an AdaBoost \cite{adaboost}, which in turn selects the discriminative appearances. This work is orthogonal to our proposal since it could help our approach to select the discriminative features while defining the appearance graph(s). 

In \cite{anton_continuous_minimization, discrete_continuous}, the authors define a mutual exclusion term based on the physical distance between two detections that occur at the same time. The term goes to infinity as the distance goes to zero. This is motivated by the fact that two objects cannot occupy the same space simultaneously. Our formulation is different in that our mutual exclusion term is defined in terms of the similarity in the label distribution rather than the position.}

\textbf{Distributed proximal optimization:} 
Our label propagation method by node-wise optimization cannot be truly characterized as a distributed computation but it raises this possibility for future developments. In such a scenario, we noticed that in \cite{chen2012fast}, the authors devise a proximal optimization on graph that has quadratic convergence by using the Nesterov's method \cite{nesterov2004convexopt}. Knowing if their approach, which assumes positive graph weights for forcing convex optimization, can be adapted to general weights and DC minimization is a matter of future study.

\textbf{Laplacian eigenmaps latent variable model (LELVM):} LELVM \cite{carreira2007laplacian} defines an out-of-sample mapping of the Laplacian eigenmaps. Given a graph, in which the weight of an edge $\bs x_i \sim \bs x_j$ is constructed as $W_{ij}:=\exp(-\|\bs x_i-\bs x_j\|_2^2/\sigma^2)$, the latent points $\bs Y$ are the solution of
\begin{align*}
	\begin{array}{ll}
		\text{minimize} & \Tr(\bs Y^\top \bs L \bs Y) \\
		\text{subject to} &   \bs Y \in \mathbb{R}^{N \times L}, \bs Y^\top \bs D \bs Y = \bs I, \bs Y^\top \bs D \bs 1=\bs 0.	
	\end{array}
\end{align*}
where $\bs D$ is a diagonal matrix with its $i$-th diagonal element defined as $D_{ii}:=\sum_j W_{ij}$, and $\bs L:=\bs D-\bs W$ is the graph Laplacian. When a new sample $\bs x$ arrives, \cite{carreira2007laplacian} defines an out-of-sample mapping $F(\bs x)=\bs y$ for a new point $\bs x$ as a semi-supervised learning problem, by recomputing the embedding as in previous equation (\emph{i.e.}, augmenting the graph Laplacian with the new point), but keeping the old embedding fixed. 
LELVM has been used for tracking human pose in \cite{lu2007people}. Our incremental label propagation is similar to LELVM in the sense that we also augment our graph and then solve for the ``latent'' label distribution. However, LELVM cannot handle newly occurring targets as it assumes that the new sample $\bs x$ belongs to one of the classes defined by $\bs X$. Moreover, it keeps the old ``latent'' distributions unchanged, which is not the case in our approach.

\section{Evaluation}
\label{section:evaluation}
The proposed algorithm has been evaluated on the following well-known and challenging datasets: APIDIS \cite{apidis}, PETS-2009 S2/L1 \cite{pets}, \tuds \cite{tud} and \tudc \cite{motchallenge}. APIDIS is a multi-camera sequence acquired during a basketball game, whereas the other three are monocular sequences.

In the remainder of the section, we first describe these datasets. We then discuss the evaluation metrics and the implementation details. Finally, we present our results and compare them with several state-of-the-art methods.
\vspace{-2mm}
\subsection{Datasets}
\label{section:datasets}
\textbf{APIDIS dataset}. 
This 1-minute video dataset is generated by 7 cameras, distributed around a basketball court. The candidate detections are computed independently at each time instant based on a ground occupancy map, as described in  \cite{detection_damien}. For each detection, the jersey color and its digit are computed to define the appearance features. In short, the jersey color is computed as the average blue component divided by the sum of average red and green components, over the foreground silhouette of the player within the detected rectangular box. The digit feature is obtained by running a digit-recognition algorithm \cite{verleysen2012recognition} in the same rectangular region. The digit feature is inherently sporadic as it is available only when the digit on the jersey faces the camera. 

\textbf{Pedestrian datasets}. To evaluate the performance of our method in monocular views, we use publicly available PETS-2009 S2/L1, \tuds and \tudc datasets. The PETS dataset is 795-frames long, with moderate target density. However, the pedestrians wear similar dark clothes, which makes appearance comparison very challenging. \revise{\tuds and \tudc are 179 and 201 frames long respectively but the  targets frequently occlude each other because of the low view-point.} 
Detection results and the ground-truth are obtained from \cite{anton_website}. Afterwards, 8-bin CIE-LAB color histograms are computed for each channel of each bounding box, resulting in a 24-bin appearance vector. We ignore the histogram(s) if the overlap ratio between any two bounding boxes exceeds 5\%. \revise{This is done because the histograms are less likely to represent the target color correctly in case of overlap, and might thus lead to wrong associations between the detections. Since the histogram feature is not available for every detection any more, it becomes sporadic.}

\subsection{Evaluation metrics}
\label{section:evaluation_metrics}
\revise{We use CLEAR MOT metric\cite{clear_mot} to evaluate our approach. It defines two quantities namely multiple object tracking precision (MOTP) and multiple object tracking accuracy (MOTA). 

MOTP is defined as the total error in estimated position for matched\footnote{A tracker output and the ground-truth are defined to be matched if their intersection-over-union ratio exceeds 50\% (respectively, if the distance $<30$ cm for APIDIS. The threshold value of 30 cm is recommended for APIDIS dataset.).} ground-truth and track pairs over all frames, averaged by the total number of matches. MOTA measures the number of misses, false positives, re-initializations and identity switches. A miss means that the tracker does not have a matching estimate for a ground-truth. Similarly, a tracker output is called a false positive when no matching ground truth is available. A switching error occurs when the tracker starts following another object, whereas a re-initialization error occurs when the tracker fails to track the object at same time and a new track is assigned for the same object later. The error due to switching is more problematic as it might lead to significant errors in higher level interpretation. 

Usually, MOTA is often preferred over MOTP because MOTP depends on the accuracy of target detector and on the accuracy of the ground-truth annotations. In our table, due to its importance regarding long term tracking capabilities, the number of switching errors (SW) is also reported.
}
\vspace{-5mm}
\subsection{Implementation details}
\label{section:implementation}
Both the joint and node-wise label propagation algorithms have been implemented on MATLAB running on a 2.4 GHz quad core CPU with 4 GB RAM. The parallel implementation of the node-wise label propagation has been done separately in C++ using Boost Graph Library and OpenMP.

\textbf{Pedestrian datasets.} For these datasets, a node is assigned to each individual detection. The size of the temporal neighborhood in spatio-temporal graph is chosen to be 10 frames. Thus, $T=10$. 
When processing time is an issue, we can envision processing the dataset in batches or running a low-level but reliable tracker first to reduce the complexity (which we perform in the APIDIS dataset).

\textbf{APIDIS dataset.} We first pre-process the data by aggregating some of the detections into tracklets based on a spatio-temporally local but reliable tracker. 
\revise{The local but reliable tracker associates two detections between successive frames into a tracklet when they are separated by less than 15 cm and there is no other detection that is closer than 15 cm from any of them.}
The resulting tracklets define the nodes in our graphs. 
The neighborhood of the spatio-temporal graph is defined to \revise{connect the tracklets} within 100 frames on each side, which allows us to connect tracklets that are up to 4 seconds apart. In the exclusion graph, the neighborhood of a node consists of all the nodes that overlap in time. 
Finally, the appearance features of a tracklet is inferred by averaging the appearance features of the detections along the tracklet.    

\revise{\textbf{Post processing.} Once the label propagation step is over, we filter out some tracks that satisfy one of the following criteria:
\begin{itemize}
	\item the number of detections along the track is less than 10 frames,
	\item the track is primarily composed of low confidence detections, \ie, if the maximum confidence value along the track is less than 0.8. 
\end{itemize}
The reasons behind these heuristics are that false tracks are usually shorter than regular target tracks and that the false positive detections have lower confidence values, compared to the true detections. This case is prevalent in PETS and both TUD datasets.
}
A glimpse of running times is presented below: 
	\resizebox{\linewidth}{!}{
	\begin{tabular}{lcccc}
		\multirow{3}{*}{} & \multicolumn{4}{c}{\textbf{Time taken}} \\ 
		\cline{2-5} 
		 & \textbf{Low-level} & \textbf{Graph} & \multicolumn{2}{c}{\textbf{Label propagation}} \\ 
		\cline{4-5} 
		\textbf{Dataset} & \textbf{tracker}  & \textbf{construction} & \textbf{Joint} & \textbf{Nodewise}  \\
		\hline 
		\tuds & - & 2 min & 3 min & 25 sec \\ 
		\tudc & - & 155 sec & 167 sec & 31 sec\\
		PETS & - & 3 min & 40 min & 5 min \\ 		 
		APIDIS & 15 sec & 1 min & 5 min & 1 min \\ 
		\hline 
	\end{tabular} 
	}
\subsection{Results}
\label{section:results}
In this section, we first present the tracking results for our frameworks, applied to offline-constructed graphs. Then, we present the tracking results for the incremental graph construction and label propagation. The computational advantages due to the node-wise decomposition  and parallelization are presented afterwards. Then, effects of parameters are discussed. Lastly, some qualitative results are presented.
\vspace{-2mm}
\subsubsection{Tracking results for offline-constructed graphs}
\label{section:results_offline}
To better compare with the literature, we consider two versions of the method. The first one uses only the spatio-temporal information. Thus, we construct only the spatio-temporal and the exclusion graphs. This is equivalent to setting $\alpha_0=1$ and $\alpha_p =0, \forall p \neq 0$ in our algorithm. In contrast, the second one considers both the spatio-temporal and the appearance features. For the \tuds, \tudc and PETS datasets, we use $\alpha_0 > \alpha_1$ ($\alpha_0$ for the spatio-temporal graph and $\alpha_1$ for the appearance graph). This constrains the spatio-temporal consistency more strictly than the appearance consistency. The reason is that  the targets wear similar clothes and therefore have similar appearances in the datasets. In the experiments, we use $\alpha_0=1$ and $\alpha_1=0.5$.\footnote{We varied $\alpha_1 \in [0.1,1]$ but did not observe significant performance changes.}

\begin{table}[h]
\centering
\def\dc{Discrete-continuous (D-C) \cite{discrete_continuous}}
\def\ce{Continuous energy \cite{anton_continuous_minimization}}
\def\gmcp{GMCP tracker \cite{gmcp_tracker}}
\def\ksp{K-shortest paths \cite{berclaz2011ksp}}
\def\gac{Global appearance (GA) \cite{horesh2011global}}
\def\iht{Iterative hypothesis (IH)\cite{kc2012iterative}}

\resizebox{0.99\linewidth}{!}{
\begin{tabular}{clccc}
{}                         & {\bf Method}                       & \multicolumn{1}{l}{{\bf MOTA}} & \multicolumn{1}{l}{{\bf MOTP}} & \multicolumn{1}{l}{{\bf SW}} \\
\myline
\multirow{7}{*}{\rotatebox[origin=c]{90}{\bf \tuds}} & \ce                 & 60.5                           & 65.8                           & 7                            \\
                                      & \dc         & 61.8                           & 63.2                           & 4                            \\
                                      & \gmcp                     & 77.7                           & 63.4                           & \bf 0                            \\
                                      \cline{2-5}
                                      & Joint (no appearance)              & 62.7                           & 73.5                           & 17                           \\
                                      & Joint (with appearance)            & 79.2                           & \bf 73.9                           & 4                            \\
                                      & Node-wise (no appearance)          & 63.1                           & 73.6                           & 16                           \\
                                      & Node-wise (with appearance)        & \bf 79.5                           & \bf 73.9                           & 4                            \\
                                      \myline
                                      
\multirow{7}{*}{\rotatebox[origin=c]{90}{\bf \tudc}} &   \dc & 57.3 & 73.7 & 13\\
& \ce & 61.6 & 73.2 & 28\\
& \gmcp & \textbf{91.63} & \textbf{75.6} & \textbf{0}\\
\cline{2-5}
& Joint (no appearance) & 62.5 & 74.3 & 12 \\
& Joint (with appearance) & 65.4 & 75.4 & 8\\
& Node-wise(no appearance) & 62.3 & 74.3 & 13\\
& Node-wise(with appearance) & 65.4 & 75.2 & 8\\
\myline                                    
                                      
\multirow{10}{*}{\rotatebox[origin=c]{90}{\bf PETS}}          & \dc         & 89.30                          & 56.40                          & -                            \\
                                      & \ce                & 81.84                          & 73.93                          & 15                           \\
                                      & \ksp                  & 80.00                          & 58.00                          & 28                           \\
                                      & \gmcp                      & 90.30                          & 69.02                          & 8                            \\
                                      & \gac            & 81.46                          & 58.38                          & 19                           \\
                                      & \iht          & 83.0                           & \bf 74.0                           & N/A                            \\
                                      \cline{2-5}
                                      & Joint (no appearance)              & 82.77                          & 71.21                          & 25                           \\
                                      & Joint (with appearance)            & 91.04                          & 70.99                          & \bf 5                            \\
                                      & Node-wise (no appearance)          & 83.07                           & 71.23                          & 25                           \\
                                      & Node-wise (with appearance)        & \bf 91.04                          & 71.00                          & \bf 5                            \\
                                      \myline
\end{tabular}
}
\caption{Tracking results on the \revise{\tuds (179 frames), \tudc (201 frames) and PETS 2009-S2/L1 (795 frames)} datasets. The D-C, IH, GMCP, KSP and GA results are obtained from \cite{discrete_continuous, kc2012iterative, gmcp_tracker, horesh2011global}.}
\label{table_offline_tud_pets}
\vspace{-5mm}
\end{table}

\revise{We compare our results with several methods such as the continuous energy (CE) minimization \cite{anton_continuous_minimization}, the discrete-continuous (D-C) minimization \cite{discrete_continuous}, the GMCP tracker \cite{gmcp_tracker}, the $K$-shortest paths (KSP) \cite{berclaz2011ksp}, the global appearance constraints (GA) \cite{horesh2011global} and the iterative hypothesis testing (IH) \cite{kc2012iterative}. The CE and D-C trackers estimate the most probable trajectories by minimizing their energies that consist in a combination of observation energy, dynamic energy, mutual exclusion energy, track persistence energy, etc. In addition, the D-C tracker uses cubic splines for modeling the motion of the target, and favors the reduction of the number of trajectories. GMCP solves greedily a generalized minimum clique problem to extract tracklets that have the most stable appearance features and the most consistent motion. KSP solves a network-flow formulation of the tracking problem and minimizes the sum of pairwise association costs between consecutive detections to estimate $K$ tracks. GA improves KSP by incorporating appearance information. IH embeds an hypothesis testing strategy into a greedy shortest-path computation procedure to exploit the appearance features that are unreliable and/or sporadically available. Since C-E, DC and KSP trackers do not use appearance information, we compare them with the first version of our approach that does not use appearance features. Similarly, since GA, IH and GMCP exploit the appearance features, we compare them to the second version of our approach.

In Table~\ref{table_offline_tud_pets}, we first observe that the joint and node-wise label optimization approaches give similar performances. For \tuds dataset, our method is better than previous methods both in terms of MOTP and MOTA. This is because our approach is able to connect the detections even if they are far in time, resulting in longer and consistent tracks. However, our method is slightly worse than GMCP in terms of ID switches. This might be because GMCP uses motion information in a global manner to ensure a smooth displacement while connecting the tracklets, which is not the case in our formalism. 

\revise{In case of \tudc dataset, our method outperforms CE and D-C. Surprisingly, GMCP has reported outstanding results. GMCP does not describe how the detections have been obtained. Our methods use same detections than CE and D-C, which has been obtained from the MOTChallenge \cite{motchallenge}}. \mycolor{black}{We have observed that removing the unreliable (confidence $<$ 0.6) detections from MOT challenge already reduces the MOTA of an oracle tracker to around 0.7, while keeping all the detections introduces many false positives. Hence, even if we were unable to run the GMCP code to verify it, we suspect that the results reported by GMCP are based on a better set of detections than the MOTChallenge ones. It is explained in detail in the supplementary material.}
 
In case of PETS dataset, again we observe that our proposed approach outperforms most approaches.  When the appearance features are ignored, the MOTA metric is better than KSP but worse than D-C. This might be because of the fact that D-C exploits higher-order motion models, whereas our formalism does not. We assert the fact that a linear motion is implicit in our formalism to justify our superior performance against KSP and GA, which do not take the motion information into account.  When the appearance information is incorporated, the performance is improved significantly from 82\% to 91\%. Moreover, the switching error is drastically reduced. 
}

The results for the APIDIS dataset are presented in Table~\ref{table_offline_apidis}. Since GA and IH are the only methods from the literature that are able to exploit sporadic appearnace features, we focus the comparison with them. As before, first we computed the results without using any appearances. This is done by setting $\alpha_0=1, \alpha_1=0, \alpha_2=0$, where the indices 0, 1 and 2 correspond to the spatio-temporal, the color and the digit graphs respectively. Afterwards, we use both the digit and the color features. As the color feature is less discriminant (because the players from the same team wear jersey of the same color) than the digit feature, we set $\alpha_1 < \alpha_2 $. Empirically, we use $\alpha_0=1, \alpha_1=0.1, \alpha_2=0.5$.

 \begin{table}[h]
	\small
	\begin{tabular}{lccc}
		\setlength{\tabcolsep}{0pt}
			\tbf{Method} & \tbf{MOTA} & \tbf{MOTP} & \tbf{SW} \\
			\hline
			IH (no appearance) \cite{kc2012iterative} & 85.83 & 60.83 & 18 \\
			IH (color+digit) \cite{kc2012iterative} & \textbf{86.19} & \textbf{60.90} & \textbf{12} \\
			GA (no appearance) \cite{horesh2011global}$^*$ & {72.91} & {53.13} & {108} \\
			GA (color+digit) \cite{horesh2011global}$^*$ & {73.07} & {53.15} & {110} \\
			\hline
			{{Joint (no appearance)}} & {81.27} & {57.13}  & {49}\\
			{{Joint (color+digit)}} & {83.90} & {60.04}  & {45}\\
			{{Node-wise (no appearance)}} & {81.4} & {57.17}  & {49}\\
			{{Node-wise (color+digit)}} & {83.85} & {60.01}  & {45}\\
			\hline
	\end{tabular}
	\caption{Results on the APIDIS dataset \revise{(1500 frames)}. The tracking results for IH and GA are been provided by the authors. [*] Since the detection results for \cite{horesh2011global} are different than that for the \cite{kc2012iterative} and ours, we relax the distance threshold to 40 cm (from 30 cm) for the tracking results of \cite{horesh2011global}. \revise{Detailed results are provided in the supplementary material.}}
	\label{table_offline_apidis}
	\vspace{-5mm}
\end{table}
Although our approach performs significantly better than GA, the results are slightly worse than IH. We see two potential reasons for this. First, our graph construction method assumes that the features are always reliable (whenever they are present). This is not the case for the IH that takes into account the confidence of feature measurement while connecting two nodes. Doing so, it lowers the impact of noisy appearance features as compared to the reliable ones. Second, IH associates two nodes only when the connection is sufficiently reliable than alternative connections. This prevents potential track switches, as reflected by the switching errors.
\vspace{-2mm}
\subsubsection{Tracking results for incrementally constructed graphs}
\label{section:results_incremental}
We constructed the graph as described in Section~\ref{section:incremental_graph_construction} and performed incremental label propagation. 
The construction of the graph in case of APIDIS dataset is slightly different than the other two datasets. In this case, if new detections can be unambiguously matched to the existing nodes, they are aggregated into a single tracklet. Otherwise, we create new nodes for the detections and connect them with existing nodes. The tracking results are presented in Table~\ref{table:incremental_results}.  
We observe that the tracking accuracy of the incremental approach is slightly worse than the off-line method. This reveals the importance of embedding a linear motion model during graph construction.
\vspace{-2mm}
\begin{table}[h]
\centering
\resizebox{0.99\linewidth}{!}{
\begin{tabular}{ccccc}
{\bf Dataset}                 & {\bf Appearance feature} & {\bf MOTA} & {\bf MOTP} & {\bf SW} \\
\hline
\multirow{2}{*}{{\bf PETS}}   & No                         & 79.32      & 70.70      & 26       \\
                              & Yes                        & 86.56      & 71.40      & 6        \\
                              \hline
{\bf TUD}    & No                         & 61.60      & 73.30      & 13       \\
{\bf Stadtmitte}                              & Yes                        & 77.20      & 73.40      & 2        \\
                              \hline
\revise{{\bf TUD}}    & \revise{No}                         & \revise{61.2}      & \revise{72.1}      & \revise{19}       \\
\revise{\bf Crossing}         & \revise{Yes} & \revise{63.4}      & \revise{72.3}      & \revise{12}        \\
                              \hline                              
\multirow{2}{*}{{\bf APIDIS}} & No                         & 74.40      & 54.20      & 52       \\
                              & Yes  (color+digit)                      & 80.23      & 58.45      & 47      \\
                              \hline
\end{tabular}
}
\caption{Results of the incremental graph construction and label propagation approach.}
\vspace{-5mm}
\label{table:incremental_results}
\end{table}

\begin{figure}[h]
	\centering
	\includegraphics[width=0.8\linewidth]{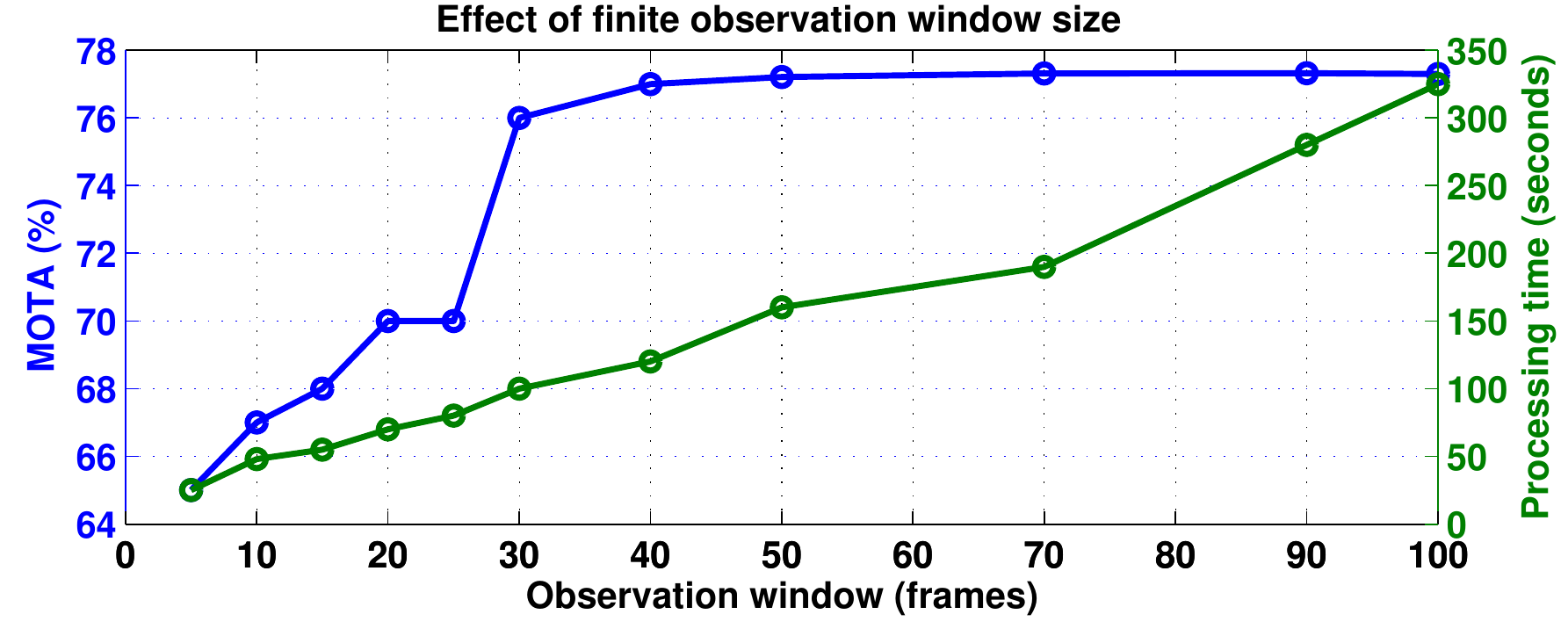}
	\caption{\textbf{Trade-off between the processing time and the
	tracking accuracy} for different observation window size for
	the \tuds dataset. {\textbestview}}
	\label{figure:incremental_to_online}
	\vspace{-2mm}
\end{figure}

To trade-off the complexity with the quality of the incremental solution, we considered only the nodes which lie within the observation window $[t-T_o,t]$ to perform label propagation. The rest of the nodes were ‘frozen’, meaning that the node-wise optimization was not
performed on those nodes. The results are elucidated in
Figure~\ref{figure:incremental_to_online} for the \tuds dataset. As we can see,
the processing time monotonically increases with the size
of the observation window. However, the tracking accuracy is improved only upto some value (50 frames in Figure \ref{figure:incremental_to_online}) after which it saturates. Alternatively, one could define other heuristic to freeze the nodes. For example, one could decide to freeze
a node if the change in its label distribution over time is
smaller than some pre-defined threshold.

\vspace{-2mm}
\subsubsection{Computational advantages of the node-wise decomposition and parallelization}
\begin{figure}[h]
	\centering
	\includegraphics[width=\linewidth]{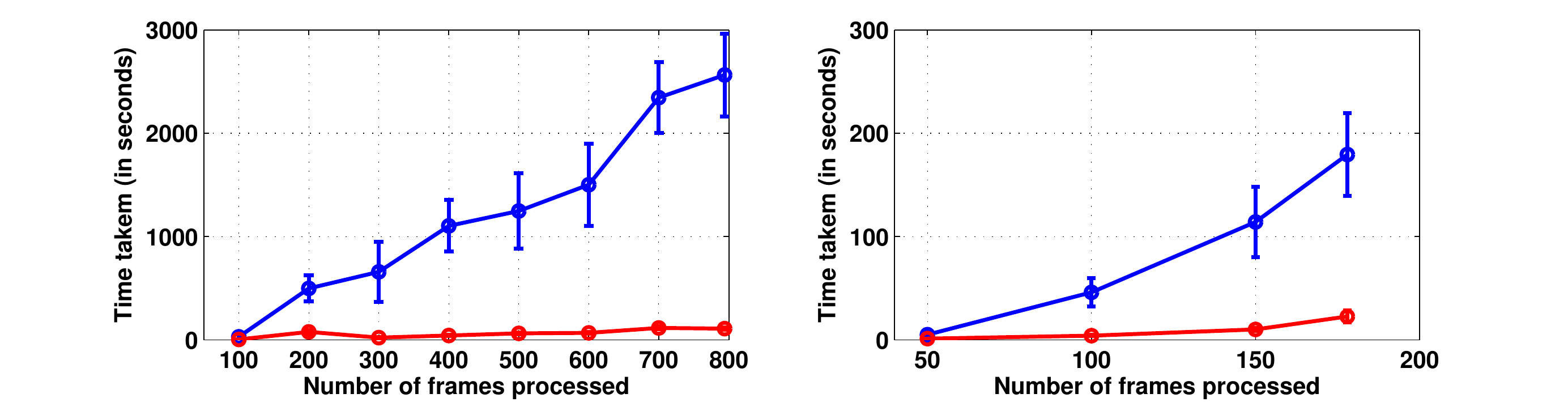} 
	\caption{Processing times for the joint and the node-wise approaches for different size of the graph. {\textbestview}}
	\vspace{-2mm}
	\label{fig:processing_time}
\end{figure}

To study the effect of node-wise decomposition, we constructed the graph off-line with different number of frames. Once the graph was constructed, we used both joint and node-wise approaches for label propagation with 10 random initializations. Afterwards, we computed the processing times for both approaches to reach the same labeling energy (equal to the labeling energy of the joint optimization after convergence). The results are shown in Figure~\ref{fig:processing_time}. We can see the dramatic improvement in computational speed, especially when the size of the graph increases. We observed that one iteration (over the whole graph) of the node-wise label optimization appears to reduce the labeling energy much faster than one iteration of the joint optimization.
\vspace{-2mm}

\begin{figure}[h]
	\centering
	\includegraphics[width=\linewidth]{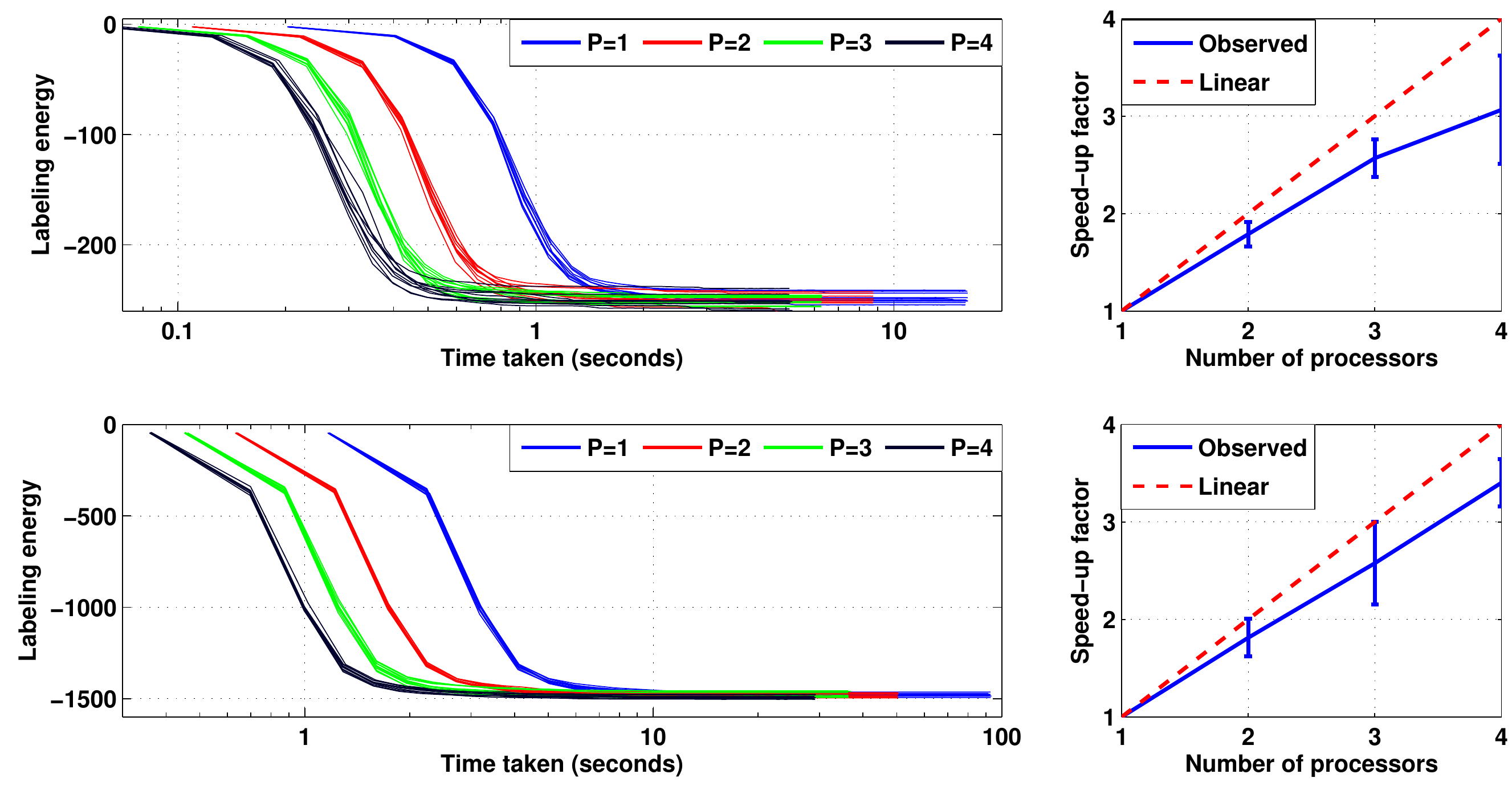}
	\caption{\textbf{Processing time and speed-up factors} for different number of processors ($P=1$ to $P=4$) of the \tuds (\textbf{top row}) and PETS (\textbf{bottom row}) datasets. For each case, we perform 10 runs of the algorithm which are drawn with the same color. \textbestview}
	\label{fig:parallel_processing_time}
	\vspace{-5mm}
\end{figure}

To assess the advantages offered by the parallel implementation, we consider a simple scheduling strategy, which directly follows the non-interference condition (see Section~\ref{section:parallel_implementation}) and selects the nodes at random. For each number of processor, we ran the algorithm 10 times and noted the evolution of objective function. The results are depicted in Figure~\ref{fig:parallel_processing_time}. The reported time is different from Figure~\ref{fig:processing_time} because of the fact that the parallel implementation is done in C++. \revise{Although the parallel implementation decreases the computational time, we observe that the reduction is not proportional to the degree of parallelism. This sub-optimal speed-up factor is due to the fact that we run the algorithm in batches of nodes. As a consequence, the time required to process a batch is governed by the longest time taken by one of its nodes. The algorithm for node-selection strategy and the distribution of time taken by nodes in the batch are presented in the supplementary material.} 
\subsubsection{Effect of parameters}
\revise{Our algorithm has some key parameters. They are listed in Table~\ref{table:table_of_notations}.
The effect of $\alpha_l$ and $T_o$ have already been discussed in Section~\ref{section:results_offline} and Section~\ref{section:results_incremental}.
In this section, we consider $T$, $T_c$, $\gamma$ and $v_{\rm max}$ and discuss 
what are their effects on the performance. For this, only one parameter is changed at a time and all other parameters are fixed at their reference values. Figure \ref{fig:effect_of_parameters} presents our results. In all graphs, the blue and green curves depicts the MOTA and the computational time respectively. In the first column, which considers the incremental algorithm, this computational time reflects both the graph construction and the label propagation, since they occur jointly all along the process. In the three last columns, which refer to the off-line algorithm, the green curve measures the graph construction time only.

\begin{figure*}
	\centering
	\includegraphics[width=0.9\linewidth]{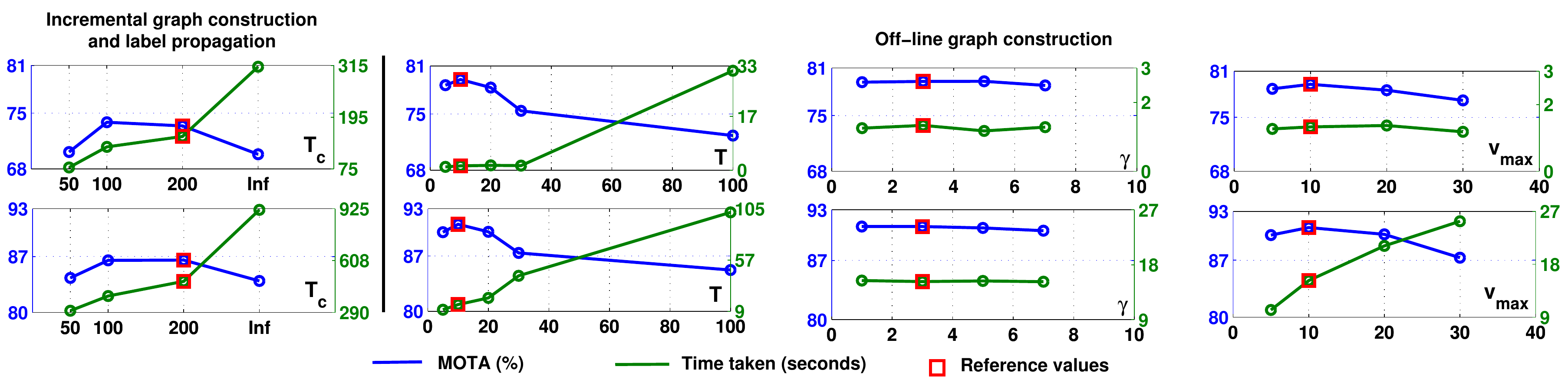}
	\caption{\textbf{Effect of parameters on \tuds (top) and PETS (bottom) datasets.} Each plot shows the effect of changing a single parameter while keeping other parameters fixed. Red squares correspond to the reference parameter values used in our experiments. The parameter is mentioned on the bottom right corner. \textbestview}
	\label{fig:effect_of_parameters}
	\vspace{-5mm}
\end{figure*}

From Figure \ref{fig:effect_of_parameters}, we observe that increasing $T_c$ increases the computation time. However, the MOTA is improved only up to some value (100 frames in our experiments) after which it starts decreasing. This is mainly due to the fact that the chances of wrong associations increase with large $T_c$. 

Since the parameters $T$, $\gamma$ and $v_{\rm max}$ do not affect the construction of the appearance graph, we report the time taken for the spatio-temporal graph only. We observe that increasing $T$ increases the connectivity of the graph (which leads to increased time to construct the graph). We observe that the MOTA increases up to certain value of $T$ and then starts decreasing again. On the one hand, when $T$ is small, it might not be effective to bridge the local missed detections. On the other hand, a large $T$ is not only more prone to wrong connections but also might not satisfy the linear motion model assumption. Interestingly, $\gamma$ does not seem to affect MOTA much.
From Figure \ref{fig:effect_of_parameters}, we also observe that $\gamma$ does not affect the graph construction time when a small window $T=10$ is considered. However, we have observed that its effect is significant when $T$ increases. As an example, the graph construction time for $\gamma=1$ is around 10 times more than that for $\gamma=7$ when $T$ is set to 100 frames for \tuds dataset.\footnote{This observation is not reported in Figure \ref{fig:effect_of_parameters}.} This is because a large $\gamma$ reinforces the implicit linear motion assumption embedded in Equation (\ref{eqn:graph_construction}), which in turn restricts the number of neighboring nodes that remain eligible for non-zero weights, leading to sharp reduction in the graph construction time.
Finally, reducing $v_{\rm max}$ typically reduces the time to construct the graph as it discards many detections that violate the gating constraint from the neighborhood. On the flip side, these detections receive non-zero weights in the exclusion graph and they receive different labels, resulting in reduced MOTA when $v_{\rm max}$ becomes too small, \ie, typically below the reference value of 10. When $v_{\rm max}$ increases beyond the reference point (in red), it increases the chances of wrong associations, resulting in lower MOTA.
}
\vspace{-2mm}
\subsubsection{Qualitative results}
\begin{figure*}
	\centering
	\includegraphics[width=0.8\linewidth]{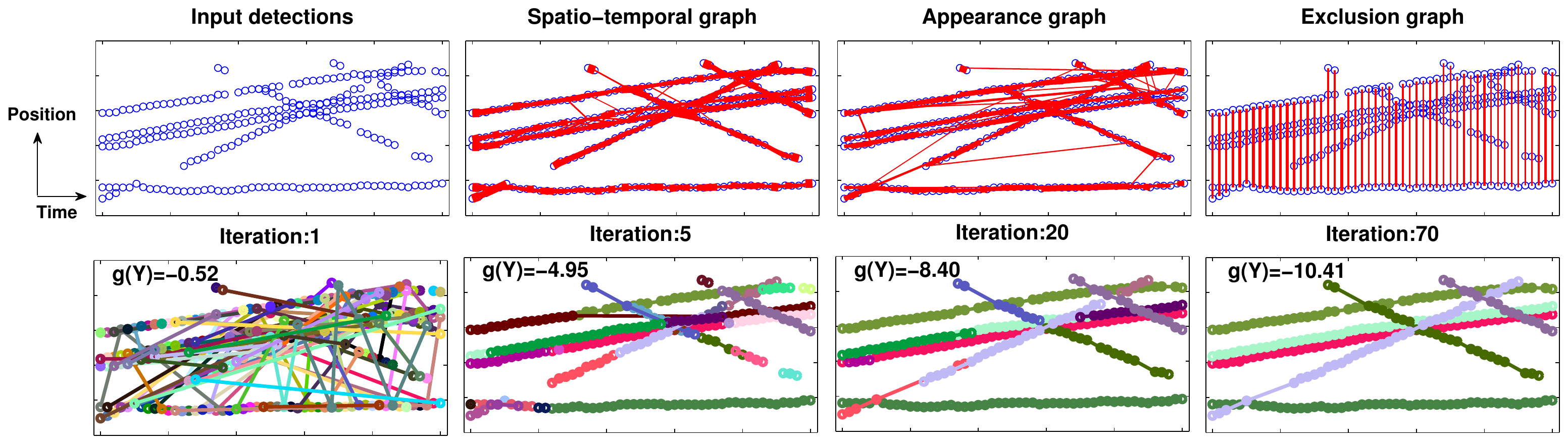} 
	\vspace{-2mm}
	\caption{\textbf{Sample graphs and label evolution on a subset of detections from PETS dataset.} Top row shows the input detections and the three constructed graphs. For clarity, edges that have weights smaller than $10^{-2}$ are suppressed. Bottom row depicts the evolution of label of the nodes along with the corresponding labeling energy. {\textbestview}}
	\label{fig:label_evolution}
	\vspace{-5mm}
\end{figure*}

Now, we present some qualitative results. Figure~\ref{fig:label_evolution} depicts the detections, constructed graphs and the inferred labels. \revise{Due to lack of space, we present the sample frames and discuss the failure cases in the supplementary material.}
\vspace{-3mm}
\section{Conclusion and future works}
\label{section:conclusion}
In this paper, we have focused on the multi-object tracking (MOT) problem under sporadic appearance features. For this purpose, a number of complementary graphs have been constructed to capture the spatio-temporal and the appearance information. Afterwards, MOT has been formulated as a consistent labeling problem in the associated graphs. The proposed solution is based on \emph{difference of convex} programming, for which we have provided both the joint as well as node-wise label optimization solutions. We show that node-wise label propagation allows us to scale up the algorithm with the number of nodes. Two further extensions of the proposed approach have been investigated. First, we have proposed a parallel implementation of the node-wise label propagation. Second, the node-wise decomposition has been embedded in an incremental graph construction step. 

Interesting paths to investigate in future research include the extensions of the framework to embed higher order motion models in the spatio-temporal graph construction, and to handle the range of features confidence levels in a continuous manner. This would be in contrast with our current approach, which turns the variable reliability of the features into sporadic measurements through hard thresholding.


\vspace{-1cm}

\begin{IEEEbiography}[{\includegraphics[width=1in,height=1.25in,clip,keepaspectratio]{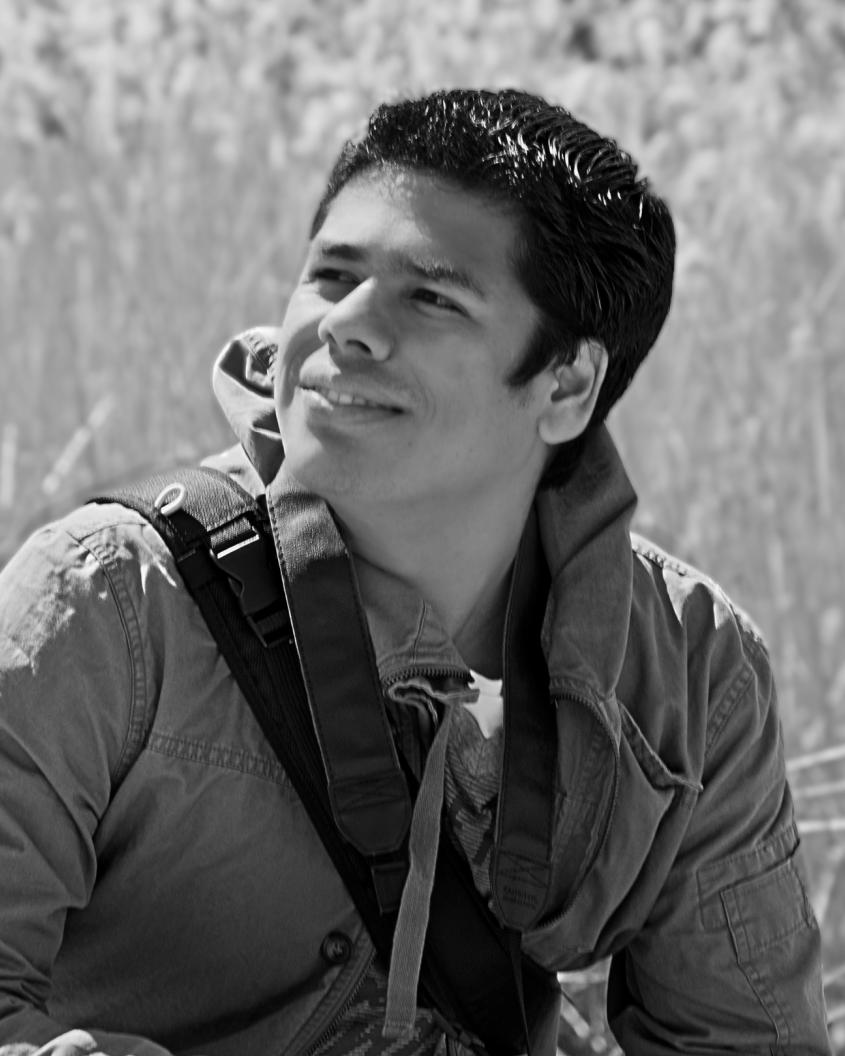}}]{Amit Kumar K.C.}
received 
double MS degree in Research on Information and Communication Technologies (MERIT) from Politecnico di Torino (PdT, Italy) and Universit\'{e} catholique de Louvain (UCL, Belgium) in 2010. Since 2010, he has been working towards his PhD in Image and Signal Processing Group (ISPGroup) in ICTEAM institute of UCL, funded by the Belgian National Science Foundation (FNRS). His research interests include multi-object tracking, graph formalism and optimization theory.
\end{IEEEbiography}

\vspace{-1cm}
\begin{IEEEbiography}[{\includegraphics[width=1in,height=1.25in,clip,keepaspectratio]{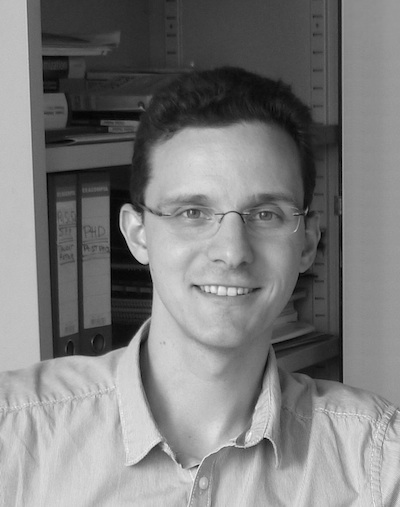}}]
{Laurent Jacques} received the BSc in Physics, the MSc and the PhD in Mathematical Physics from the Universit{\'e} catholique de Louvain (UCL), Belgium. Postdoctoral researcher in the ICTEAM institute of UCL from 2005 to 2011, he was funded by the Walloon Region (2005-2006), the Belgian FRS-FNRS (2006-2010, 2011-2012) and by the Belgian Science Policy (Return Grant, BELSPO, 2010-2011). Visiting researcher at Rice University (DSP/ECE, Houston, TX, USA) in spring 2007, he also performed a postdoctoral stay from 2007 to 2009 at the Swiss Federal Institute of Technology (LTS2/EPFL, Switzerland). Since Oct. 2012, he is Professor and FNRS Research Associate in the Image and Signal Processing Group (ISPGroup) in ICTEAM/UCL. His research focuses on Sparse Representations of signals (1-D, 2-D, sphere), Compressed Sensing theory (reconstruction, quantization) and applications, Inverse Problems in general, and Computer Vision.
\end{IEEEbiography}

\vspace{-1 cm}
\begin{IEEEbiography}[{\includegraphics[width=1in,height=1.25in,clip,keepaspectratio]{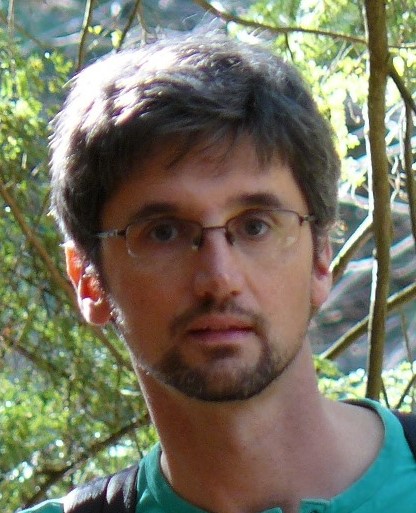}}]{Christophe De Vleeschouwer} is a Senior Research Associate at the Belgian NSF, and an Associate Professor at UCL (ISPGroup). He was a senior research engineer with IMEC (1999-2000), a post-doctoral Research Fellow at UC Berkeley (2001-2002) and EPFL (2004), and a visiting scholar at CMU (2014-2015). His main interests concern video and image processing for content management, transmission and interpretation. He is enthusiastic about non-linear and sparse signal expansion techniques, ensemble of classifiers, multi-view video processing, and graph-based formalization of vision problems. He is the co-author of more than 35 journal papers or book chapters, and holds two patents. He served as an Associate Editor for IEEE Transactions on Multimedia, has been a co-founder of Keemotion(\url{www.keemotion.com}), using video analysis for automatic sport coverage.
\end{IEEEbiography}

\end{document}